\documentclass[final,5p,times,twocolumn]{elsarticle}

\usepackage{amssymb} 
\usepackage{amsthm} 
\usepackage{graphicx} 
\usepackage{natbib} 
\usepackage{multirow} 
\usepackage{hyperref} 
\usepackage{amsmath} 

\makeatletter
\def\ps@pprintTitle{%
  \let\@oddhead\@empty
  \let\@evenhead\@empty
  \def\@oddfoot{\reset@font\hfil\thepage\hfil}
  \let\@evenfoot\@oddfoot
}
\makeatother

\begin{document}
\begin{frontmatter}

\title{Personalized Lane Change Decision Algorithm Using Deep Reinforcement Learning Approach}

\author{Daofei Li\corref{cor1}}
\cortext[cor1]{Corresponding author}
\ead{dfli@zju.edu.cn}

\author{Ao Liu}

\address{Institute of Power Machinery and Vehicular Engineering, Faculty of Engineering, Zhejiang University, No 38 Zheda Road, Xihu District, Hangzhou, 310028, Zhejiang, China.}

\begin{abstract}
To develop driving automation technologies for human, a human-centered methodology should be adopted for ensured safety and satisfactory user experience. Automated lane change decision in dense highway traffic is challenging, especially when considering the personalized preferences of different drivers. To fulfill human driver centered decision algorithm development, we carry out driver-in-the-loop experiments on a 6-Degree-of-Freedom driving simulator. Based on the analysis of the lane change data by drivers of three specific styles,personalization indicators are selected to describe the driver preferences in lane change decision. Then a deep reinforcement learning (RL) approach is applied to design human-like agents for automated lane change decision, with refined reward and loss functions to capture the driver preferences.The trained RL agents and benchmark agents are tested in a twolane highway driving scenario, and by comparing the agents with the specific drivers at the same initial states of lane change, the statistics show that the proposed algorithm can guarantee higher consistency of lane change decision preferences. The driver personalization indicators and the proposed RL-based lane change decision algorithm are promising to contribute in automated lane change system developing.
\end{abstract}

\begin{keyword}
Reinforcement learning, Deep Q-Network, Decision making, Lane change, Driver-in-the-loop, Driving style
\end{keyword}

\end{frontmatter}


\section{Introduction}\label{sec1}

Automated driving has a been a hot topic in recent technology developments in automotive and transportation industries. Notable companies in this technology competition, such as Google Waymo, Baidu, and Cruise Automation, have recently launched their top-ranking products of self-driving cars with SAE Level 4 automation, most of which are even for driverless robotaxi ride-hailing services on open roads \cite{bib1}\cite{bib2}. On the other hand, the Advanced Driving Assistance Systems (ADAS) in production cars, by enhancing safety and comfort, have been widely recognized by customers , e.g. Adaptive Cruise Control (ACC), Lane Keeping Assist (LKA), Navigate on Autopilot (NOA) by Tesla \cite{bib3}, Navigate on Pilot (NOP) by NIO \cite{bib4}.

Research on earlier-introduced ADAS functions, e.g., ACC and LKA, shows that personalization of ADAS functions is necessary to improve both driving comfort and safety \cite{bib5}. Before the fully automated driving, i.e., in SAE Level 4, is possible for mass production in the future, there is no doubt that human drivers are still necessary to sit behind the wheel to supervise or even to handle the daily driving tasks, though parts of which may be actually assisted by driving automation. Therefore, for a long time to come, a well-designed driving automation strategy should be compatible with or even aligned to human preferences, in order to better support human drivers in driving. Regardless of partial or full levels of driving automation, if the actual assistances in fulfilling driving tasks are congruent with personal styles in manual driving mode, the customer trust and acceptance of assistant functions can also be guaranteed. 

\subsection{Lane change:a dynamic and stressful driving maneuver}\label{subsec2}

Lane change is considered more complex than other driving maneuvers, drivers need to take into account several vehicles in the current lane and target lane, which increases their stress. For the lane change assistance, a fundamental reason behind personalization is that different drivers have different criteria for judging safety, e.g. the acceptable gap, relative distance and approaching rate. A universal assist design can cause a lot of problems, it may too conservative for aggressive drivers and too aggressive to conservative drivers \cite{bib6}. This may prevent the driver from activating the assist function, which is regrettable because they cannot take advantage of its benefits. But some more serious accidents may happen if the driver cannot understand the behaviors or decisions made by assistant function. For example, the assistance system makes a lane change decision and starts merging, although safety can be guaranteed, it does not match the driver's intuition, the driver may intervene and turn the steering wheel. It is conceivable that this can cause a serious accident, especially at high speeds. Actually, similar contradictions between the autopilot system and the pilot have caused serious aviation accidents. Therefore, to avoid such incidents occurring on self-driving cars in the future, personalization is a feasible way to make the driver to understand the assistant system, so as to eliminate the contradictions.

To incorporate some certain personalized driving styles in driving automation, a straightforward way of developing an automated or autonomous strategy is to imitate or even to replicate a specific driver’s driving operations in the addressed scenarios, which is actually to model the driver behaviors. A generalized framework of driver model, equipped with general intelligence and capable of solving general driving tasks, is definitely more attractive but is not available yet. In consideration of the radical variation of driving scenario condition, driver modelling is usually done in scenario-by-scenario fashion, i.e., a holistic driver model is actually an integrated model of several sub-models of driver behaviors, for example, car following, lane change, and steering, etc \cite{bib7}.

Lane change is a common but complicated maneuver in driving, for which various of driving assistance systems have been provided, e.g., Blind Spot Detection (for warning only), Lane Change Assist with Turn Assist, or even Auto Lane Change (ALC). Among of them, ALC, as provided both in Tesla NOA and NIO NOP, is in particular need to be designed in line with drivers’ preferences or personal styles of maneuvering. Even in hands-free driving mode, the driver may still experience the dynamic and stressful highway lane-changing scenarios on the edge of incidents or even accidents. Therefore, it is critical to design a safe, comfortable and personalized algorithm for automated lane change maneuvers.
A lane change process can be further divided into three sub-stages, i.e. (a) sensing, (b) decision making and planning, (c) and control execution, which is similar to the sense-plan-act methodology in robotics. All three sub-stages contribute to represent a driver’s personalized style in a lane change process, while most prominently the decision making sub-stage shows the drivers’ personalized preferences, especially in stressful lane change maneuvers under dense traffic conditions. Therefore, in this paper we focus on the personalized decision making in the context of automated lane change algorithms.

\subsection{Review of lane change decision models}\label{subsec3}

In the past decades, many scholars have carried out in-depth studies on lane change decision making of human drivers. Review papers have classified lane change decision models from different perspectives \cite{bib8}\cite{bib9}\cite{bib10}, while according to research methods, they can be summarized mainly into two groups: rule-based and learning-based models.

As for rule-based decision models, Gipps \cite{bib11} is one of the earliest scholars to study drivers’ lane change behavior, who proposed a rule-based decision making framework for the urban traffic on the basis of his car-following model \cite{bib12}. Gipps’ model summarized the lane change decision making process as a flowchart, in which traffic signals, obstacles, heavy vehicles and some other factors are taken into account. And thanks to the flexibility of the framework, any factors that affect lane change decision making can be added or replaced. Although there is no consideration of drivers’ variational behavior in it, Gipps’ model still has a profound influence on the subsequent researches on the lane change decision model. Halati et al. \cite{bib13} developed a lane change model applied in CORSIM, in which lane change behaviors are classified into mandatory lane changing (MLC), discretionary lane changing (DLC) and random lane changing (RLC). Motivation, advantage and urgency are three main factors considered in this model during a lane change process, but the execution depends on the availability of the gap between preceding vehicle and following vehicle in the target lane. Lane change models applied in the FRESIM \cite{bib14} and NETSIM \cite{bib15} are similar but different in the ways of calculating the acceptable gap. In 2007, Kesting \cite{bib16} proposed a general lane change decision making model, MOBIL, in which the IDM model \cite{bib17} is used to compare the total deceleration of all surrounding vehicles before and after lane change, and the decision will be made afterwards. A parameter called “policy factor” is considered in MOBIL to reflect the cooperation between drivers during lane change process. Additionally, in case of merging and congested scenarios, game theory is used to research the lane change decision making behavior in \cite{bib18}\cite{bib19}, but these approaches may not be easily applied to other lane change scenarios with more surrounding cars to consider. And also, the drivers’ personalized preferences are neglected.

With the development of machine learning, more and more approaches based on deep learning or deep reinforcement learning are proposed to model the lane change decision making, since the neural networks allow us to consider more factors that influence lane change decision and the self-learning feature make the algorithm more robust. Vallon et al \cite{bib20} propose a data driven modeling approach to capture the lane change decision behavior of human driver, with the help of SVM classifiers they can predict the drivers’ lane change intention without explicit initiation, and the results show that the personalized can reproduce the lane change behavior between different drivers after the classification finished by the classifiers. Mirchevska et al \cite{bib21} design an RL agent by using a Deep Q-Network, their aim is to make the agent drive as close as possible to a desired velocity. A baseline rule-based agent is tested at the same time but the trained RL agent shows better performance. Hoel et al \cite{bib22} train a Deep Q-Network agent for a truck-trailer combination in a highway driving case, they teach the RL agent to finish overtaking maneuvers by making lane change decision, the test result shows the RL agent perform better than a commonly used reference model. Machine learning allows researchers to consider more influential factors during modeling the lane change decision, but once drivers’ personalized preferences are taken into account, these data-driven approaches require a lot more of lane change data for a particular driver. 

Learning-based, or data-driven, modeling approaches are inherently data-hungry, while the level of data size, data coverage and data detail, determine the model scope and also the potential application domain. There are already some public datasets of naturalistic vehicle trajectories, such as the Next Generation SIMulation (NGSIM) program launched by the Federal Highway Administration (FHWA) \cite{bib23} and the highD dataset published by RWTH Aachen University \cite{bib24}. However, regardless of whether the data collection is via cameras mounted in hovering drones or cameras fixed on traffic sign poles, for a specifically investigated vehicle-driver combination, only a very limited range of driving trajectory and duration are available. For some lucky cases of lane change, the vehicle poses and motion data during the whole process are possible for building a statistical model of lane change behavior, which represents multiple lane change instances by multiple drivers. These open datasets are suitable for studying on some general problems of lane change model among the human drivers, but not for the personality research of a specific driver. 

Most previous researches only consider the general problem in lane change decision making and aim to capture the general behavior of driving, e.g. to make the lane change process safer and faster, or to make the traffic flow more efficient. Due to the lack of data in driver-specific level and also the model framework limits, the personalized preferences of in-vehicle drivers cannot be appropriately studied.

\begin{figure*}[t]
\centering
\includegraphics[width=1.05\textwidth]{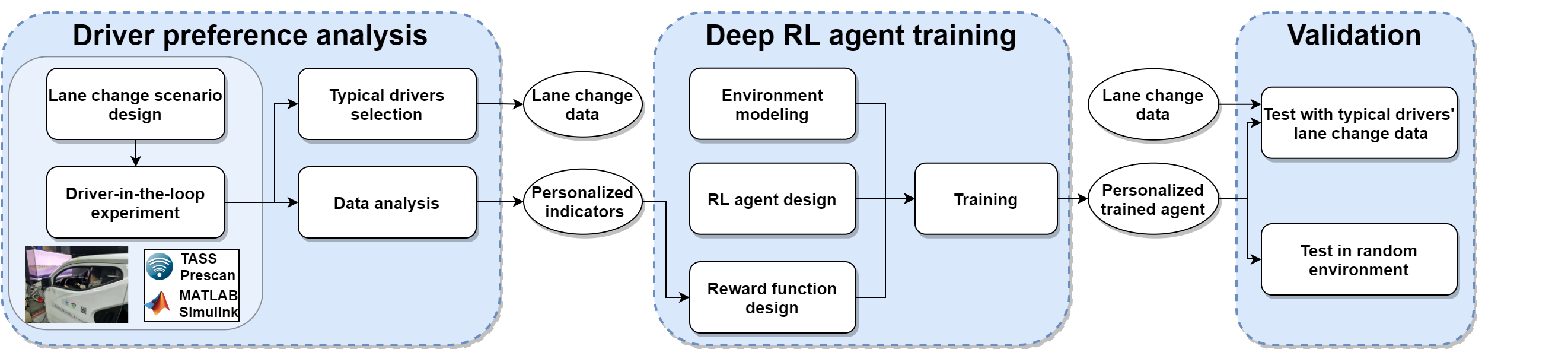}
\caption{The proposed algorithm structure of personalized lane change decision}\label{fig1}
\end{figure*}

In this paper, we want to consider the driver’s preference in lane change decision and to solve the data collection problem mentioned above. Our work focuses on personalized automated lane change decision making in a two-lane highway scenario. Fig.\ref{fig1} shows how we achieve the personalized lane change decision algorithm, from raw data collection, analysis, RL algorithm design, to validation. More specifically for the adopted approach, we design the RL agents to make lane change decision with perception of surrounding environment, and from the personalization reward function the agent learns how to make the lane change decision close to the specific driver’s personalization as expected. The different personalized preferences are described with three indicators obtained from drivers, which are validated by Driver-in-the-loop (DIL) experiments in a 6-Degrees-of-Freedoms (6-DoF) simulator. 

The rest of the paper is organized as follows. Section \ref{sec2} gives a statement about our lane change decision making problem and a brief introduction of the reinforcement learning. Section \ref{sec3} presents the DIL experiment results and drivers’ lane change preference analysis. The RL based lane change decision making and the training procedure are described in Section \ref{sec4}. The results of the RL agent’s training and test are summarized in Section \ref{sec5}. Finally, the conclusions and some potential future work are presented in Section \ref{sec6}.


\section{Problem statement}\label{sec2}

\begin{figure}
\centering
\includegraphics[width=0.5\textwidth]{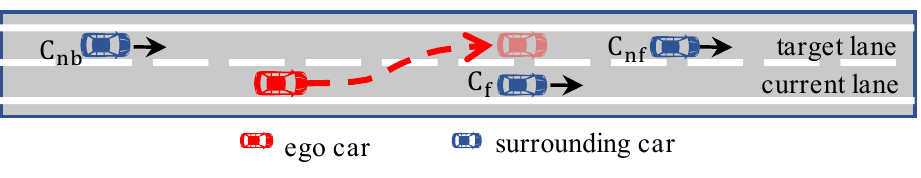}
\caption{Lane change decision scheme in a two-lane driving scenario}\label{fig2}
\end{figure}

Considering the lane change decision task as a reinforcement learning (RL) problem, we design the RL agents to make lane change decision just like a personalized driver, specifically in high-speed driving scenarios with multiple surrounding cars. Deep-Q-Network are used to learn the state-action value function for describing the reward of personality, $Q(s,a)$. We consider a two-lane highway scenario with three surrounding cars at random speeds, as shown in Fig.\ref{fig2}. 

The notations of surrounding cars are as follow: the ego car ($C_{ego}$), the front car in the current lane ($C_f$), the front car in the target newer lane ($C_{nf}$), and the behind car in the target newer lane ($C_{nb}$).  In this situation, taking into account the influences of the surrounding cars, the ego car decides whether or when to change from the current lane to the target lane.

\subsection{Reinforcement learning}\label{subsec2}

It is complex to make a lane change decision with multiple surrounding cars especially considering driving personalization. In this paper, the three personalization indicators need to be taken into account simultaneously, however, in majority of cases they cannot reach the optimum at the same time. Therefore, we need to trade off and give the overall optimal decision. On the other hand, innumerable environmental conditions and limited typical driver lane change data require our algorithm have the ability of self-learning and wide adaptability. The Reinforcement Learning approach meets our requirements.

Reinforcement learning is a branch of machine learning \cite{bib25}. In an RL problem, an agent selects an action, $a$, depends on the current state of the environment, $s$, and the policy, $\pi$, then the environment will change to a new state, $s'$, and return a reward, $r$. The goal of an RL problem is to find an optimal policy, $\pi^*$, that maximizes the cumulative reward. The cumulative reward $R_t$ defined as

\begin{equation}
R_t = \sum\limits_{k=0}^{\infty}\gamma^kr_{t+k},\label{eq1}
\end{equation}
where $r_{t+k}$ is the reward returned at $t+k$ step, and $\gamma$ is a discount factor, $\gamma \in [0,1]$.

\subsection{Q-learning and Deep-Q-Network}\label{subsec3}

State-action value function $Q(s,a)$ is used to evaluate the expected cumulative reward of agent when selecting action $a$ in state $s$, defined as
\begin{equation}
Q(s,a) = E[R_t \vert s_t=s,a_t=a].\label{eq2}
\end{equation}

In Q-learning algorithm, the optimal state-action value function can be presented as
\begin{equation}
Q^*(s,a) = E[r+ \gamma \max\limits_{a'}Q^*(s',a') \vert s_t=s, a_t=a].\label{eq3}
\end{equation}

When the $Q^* (s',a')$ is known, the optimal policy is to select an action $a'$ that maximizes the $Q^* (s',a')$.

Q-learning is a classical algorithm for the problem with limited states and actions, the state-action values are saved in a Q-table. However, if the state space is continuous, it is impractical to remember the state-action values with a table, so we need a continuous function to approximate, and the neural network is a good choice. The Deep Q-Network (DQN) algorithm \cite{bib26}\cite{bib27}, approximates the optimal state-action value $Q^* (s,a)$ with a nonlinear estimator $Q(s,a;\theta)$ based on deep neural network. Network weights $\theta$ will be updated during the training process to minimize the loss function defined as
\begin{equation}
L(\theta) = E[(r+ \gamma \max\limits_{a'}Q(s',a',\theta)-Q(s,a,\theta))^2].\label{eq4}
\end{equation}

However, the same network weights leading to an unstable training. To solve this problem, the target network weights are set to $\theta^-$ and will be replaced by the prediction network weights $\theta$ every fixed steps. And the final loss function is defined as
\begin{equation}
L(\theta) = E[(r+ \gamma \max\limits_{a'}Q(s',a',\theta^-)-Q(s,a,\theta))^2].\label{eq5}
\end{equation}


\section{Driver preferences in lane change}\label{sec3}

To model the human driver preferences in lane change decision, an ideal dataset should be from several different drivers’ naturalistic driving on public roads. However, considering lane change as one kind of highly dynamic and time-critical process, if the lane change decision timing matters, it is especially difficult to precisely record the exact decision schedule during one lane change episode. As one improved way of data collection, experimental driving on test roads can seemingly better assure the realistic driving condition and the experiment drivers can almost maneuver the vehicle as he or she usually does, if the surrounding cars (usually 2 to 3 additional cars) can be coordinated well by human or robot drivers. However, experimental driving, if with only limited time and/or limited financial budget, is still impossible to collect enough data of lane change decision strategy, not to mention that the experiment safety risk is extremely high due to the involvement of multiple vehicles at high speeds. 

Therefore, in order to analyze the drivers’ preferences in lane change decision, here we adopt a driving simulator for original data collection for decision preference analysis, and then after confirmation, three drivers with different driving styles are selected, which will be the target personalization style for RL agents.

\subsection{Driver-in-the-loop experiments}\label{subsec2}

\begin{figure}
\centering
\includegraphics[width=0.5\textwidth]{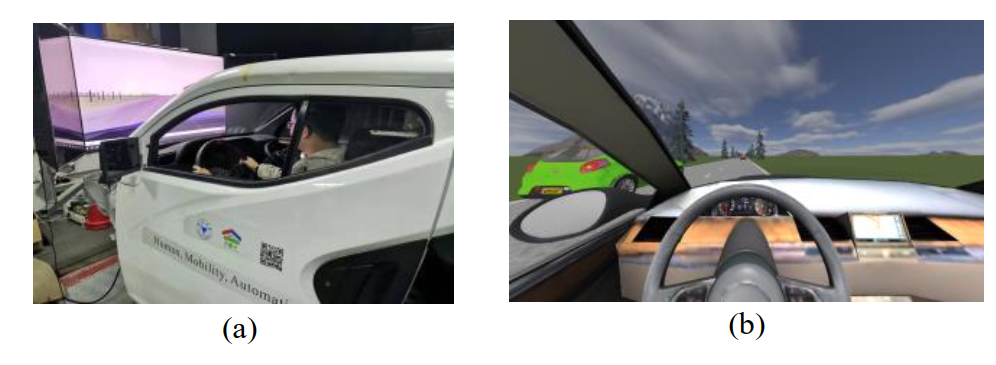}
\caption{(a) The 6-DoF driving simulator. (b) Lane change scenario in Prescan}\label{fig3}
\end{figure}

We design a driver-in-the-loop (DIL) experiment environment based on  a 6-Degrees-of-Freedom (6DoF) driving simulator, which can provide a realistic driving experience, as shown in Fig.\ref{fig3}. A two-lane highway driving scenario is established in the simulation software, TASS Prescan, while the surrounding cars with random constant speeds are controlled by the MATLAB/Simulink model, and the ego car is controlled with a real steering wheel and gas/brake pedal inputs given by human drivers. A specific button on the steering wheel is set for recording the time stamp of every lane change initiation, while the rest of collected data throughout the experiments include the position and speed of all vehicles, the driver inputs, etc.

Ten drivers are invited to participate in our DIL experiment, aged between 20 to 26, each driver is asked to make the lane change maneuvers for 50 times on four sections with different speed limits, 60 kph, 70kph, 80kph and 90kph. Due to the fact that not all drivers can exactly follow the speed limits, a maximum 5 kph speed error over limits are still considered effective.

\begin{figure}
\centering
\includegraphics[width=0.5\textwidth]{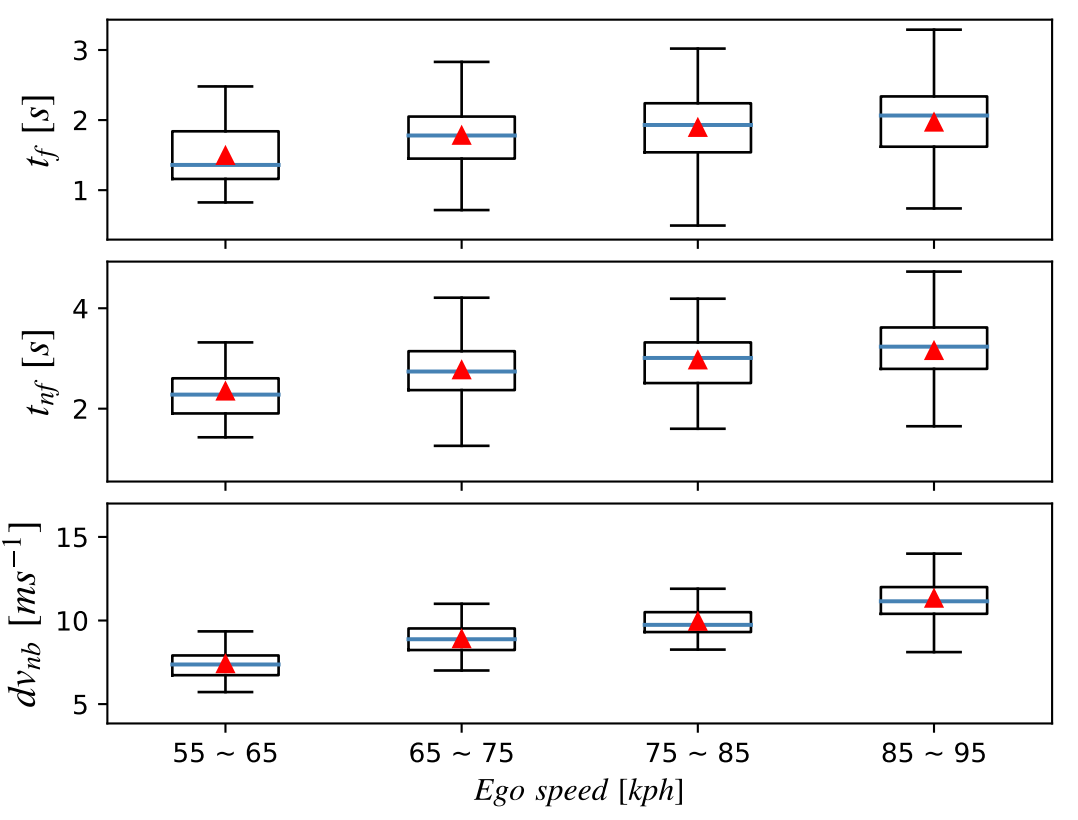}
\caption{The statistical results of ten human drivers’ lane change decision data in DIL experiment in different speed ranges. (blue lines represent median values of indicators in different speed ranges; red triangles represent mean values.)}\label{fig4}
\end{figure}

According to the existing researches on natural driving data \cite{bib28}\cite{bib29}, there are many factors affecting drivers' lane change decision, such as relative velocity, time to collision (TTC) and relative distance. In this paper, drivers should consider avoiding collisions with surrounding cars. For the front car in current lane and the target lane, drivers mainly judges whether a collision will occur by sensing the relative speed and distance, which can be described by the value of TTC, $t_f$ and $t_{nf}$, respectively. And as for the behind car in target lane, the approaching rate, which can be described by the relative speed, $dv_{nb}$, is a more direct factor of judgement, because the perception of relative distance is not intuitive. Therefore, drivers’ different preference in lane change decision corresponds to a different combination of values of the three potential personalization indicators, i.e., $t_f$, $t_{nf}$ and $dv_{nb}$. After analyzing the lane change data of drivers, we find that these three indicators are  almost correlated positively to the velocity of ego car, $v_e$. Fig.\ref{fig4} shows these ten drivers' statistical results of three indicators in different speed ranges. Further, correlation analysis is used to ensure these variables'  relationship  for all the drivers, and the results reveal that 80\% of the drivers show a significant linear correlation  between $v_e$ and three indicators. Therefore, we define the driver personalization by three indicators, i.e. the indicator set defined as

\begin{equation}
I_{dp} = [t_f, t_{nf}, dv_{nb}]^T.\label{eq6}
\end{equation}

With the above analysis, these discrete points are fitted with linear regression and the fitting curves can be described as 
\begin{equation}
I_{dp} = Av_e + b \label{eq7}
\end{equation}
which will be used as reference lines in reward function design.

\subsection{Drivers with typical personalization preferences}\label{subsec3}

\begin{figure}
\centering
\includegraphics[width=0.5\textwidth]{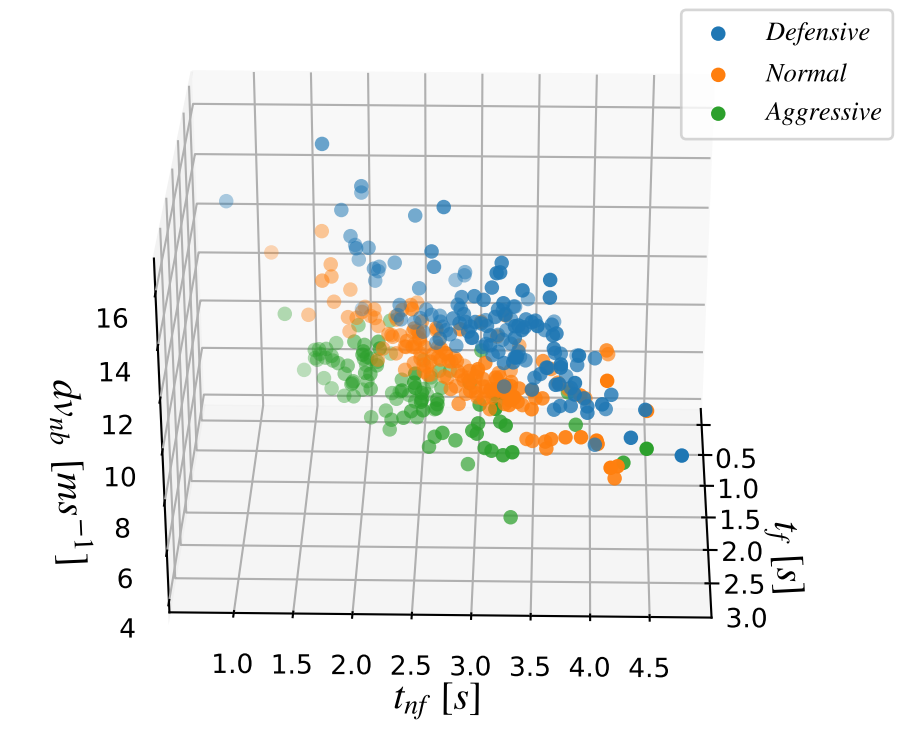}
\caption{The clustering result of three styles of driver lane changing decision}\label{fig5}
\end{figure}
\begin{figure}
\centering
\includegraphics[width=0.5\textwidth]{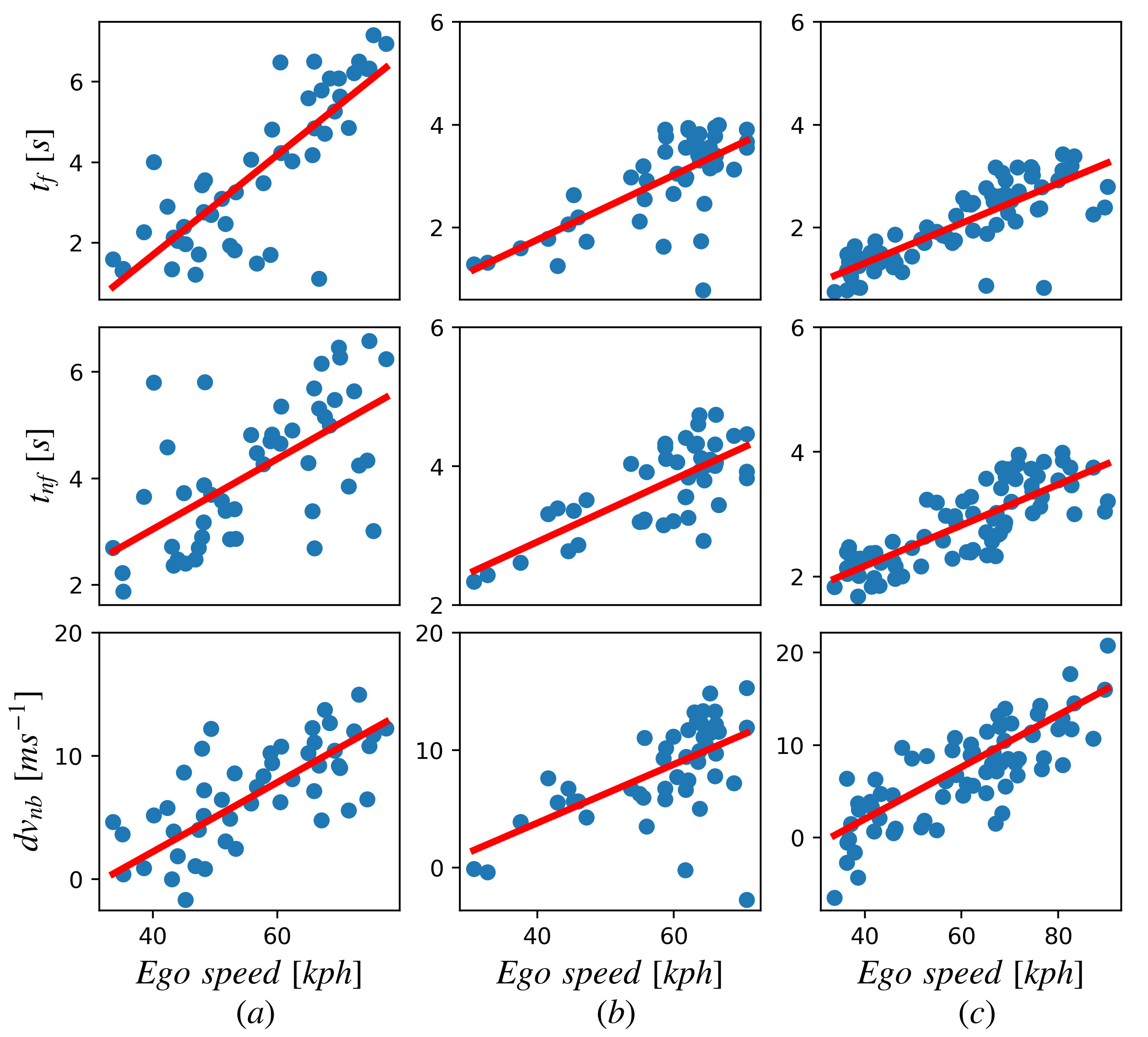}
\caption{Change of three personalization indicators with ego speed at decision point, the red lines are the regression curves obtained by ordinary least square (OLS). (a) Defensive driver. (b) Normal driver. (c) Aggressive driver.}\label{fig6}
\end{figure}

\begin{table}

\caption{Parameter matrix for different type of drivers obtained by OLS}\label{tab1}
\begin{center}
\begin{tabular}{lccc}
\hline
Type & $A$  & $b$ \\ \hline
Defensive    & $[0.45,0.24,1.01]^T$   & $[-3.26,0.42,-9.02]^T$  \\
Normal    & $[0.23,0.16,0.90]^T$   & $[-0.75,1.11,-6.18]^T$  \\
Aggressive    & $[0.14,0.12,1.01]^T$   & $[-0.25,0.86,-9.25]^T$  \\ \hline

\end{tabular}
\end{center}
\end{table}

After DIL experiments, drivers' lane change data with various personalization preferences are obtained. Assuming that there are three general personalization styles, we cluster the drivers' lane change decision into three groups denoted as Defensive, Normal and Aggressive, as show in Fig.\ref{fig5}. At the initiation of lane change decision, the more aggressive drivers, the smaller their TTCs($t_f$, $t_{nf}$) and relative speed($dv_{nb}$). 

We select three drivers with obvious characteristics differences as typical examples according to the clustering results. As shown in Fig.\ref{fig6}, for each driver, the value of the indicators,  $t_f$, $t_{nf}$ and $dv_{nb}$ increase with the speed of ego car, $v_e$, linearly, and the curve parameter matrix $A$, $b$ are obtained via ordinary least square (OLS) method and presented in Table \ref{tab1}. 


\section{RL-based lane change decison making}\label{sec4}

\begin{figure}
\centering
\includegraphics[width=0.5\textwidth]{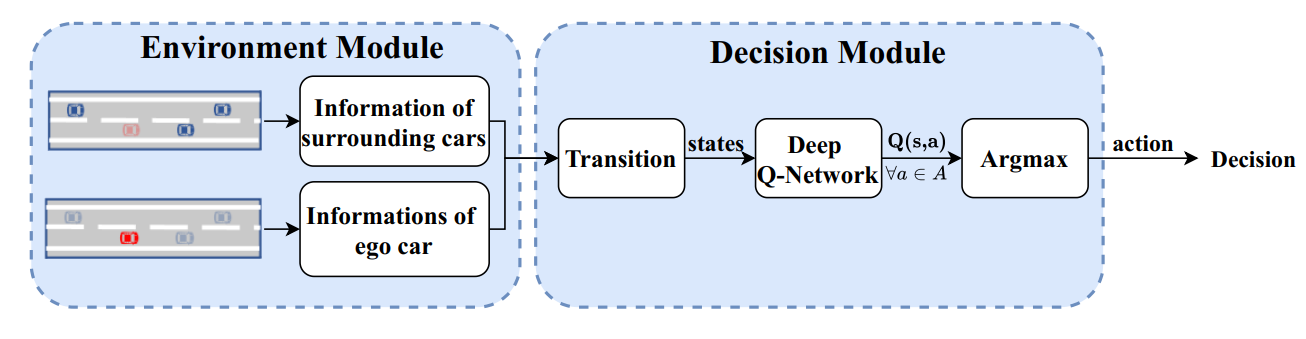}
\caption{The process of lane change decision making}\label{fig7}
\end{figure}

To formulate the lane change decision problem in RL-based framework, we define action space, state space, reward function and Deep Q-Network, to find the optimal lane change decision policy. Fig.\ref{fig7} schemes the decision making process in lane change. The RL environment module provides information of ego car and surrounding cars to the decision module, this information is transitioned to the state vector as an input of Deep Q-Network. Then the network output the evaluations of state-action value for each action, and the best action is selected as the final decision.

\subsection{Action space}\label{subsec2}

In this paper, the left lane change decision or right lane change decision are the same, they all mean that the ego car expects to move from the current lane to the target lane. So, there are only two discrete actions in our lane change decision making problem, i.e., $a_1$:TO CHANGE lane to the target lane, and $a_2$: NOT TO CHANGE lane (to keep in the current lane). The action space A is defined as

\begin{equation}
A = \{a_1, a_2\}.\label{eq8}
\end{equation}

\subsection{State space}\label{subsec3}

Based on the analysis result in Section \ref{sec3}, the personalization indicators include $t_f$, $t_{nf}$ and $dv_{nb}$, and they need to be obtained from the state of environment. The state consists of two parts, information of ego car and information of surrounding cars. Every car's information includes the longitudinal velocity, $v$, and the longitudinal position, $x$. For better performance in training, we normalize the $v$ and $x$ to $(0,1]$ respectively. Therefore, the state can be described as a vector of eight normalized continuous values

\begin{equation}
s = [v_e, x_e, v_f, x_f, v_{nf}, x_{nf}, v_{nb}, x_{nb}].\label{eq9}
\end{equation}

\subsection{Reward functions}\label{subsec4}

In order to train the human-like RL agents, the personalization indicators mentioned in Section \ref{sec3} are used for reference to design the reward functions. To help the agent trade off the benefits between different decisions and learn to make a better choice, sum of two actions' reward is kept as a constant at every decision step. The form of reward function for each indicators is designed as follows:
If the decision is TO CHANGE lane, i.e. $action=a_1$,  

\begin{equation}
r = \left\{
\begin{aligned}
&1&,&e_i \in [0,m]\\
\frac 1{m-n} * &e_i- \frac n{m-n}&,&e_i \in (m,n)\\
&0&,&e_i \in [n,+\infty) \label{eq10}
\end{aligned}
\right.
\end{equation}

If the decision is NOT TO CHANGE lane, i.e. $action=a_2$,

\begin{equation}
r = \left\{
\begin{aligned}
&1&,&e_i \in [0,m]\\
\frac 1{n-m} * &e_i- \frac m{n-m}&,&e_i \in (m,n)\\
&0&,&e_i \in [n,+\infty) \label{eq11}
\end{aligned}
\right.
\end{equation}

For each indicator $i \in I_{dp}$ in (\ref{eq6}), $e_i$ is the absolute error of actual value $i_{act}$, and reference value $i_{ref}$, which is defined as

\begin{equation}
e_i = \vert i_{act}-i_{ref} \vert.\label{eq12}
\end{equation}

The smaller the $e_i$, the more suitable for personalization lane change in terms of indicator $i$, and if the agent chooses to change lane, it will receive a greater reward than choosing not to change lane. Instead, if $e_i$ is large enough, the agent need to choose to keep in the current lane for a greater reward. 

The variables $m$ and $n$ are two preset parameters, which represent the maximum acceptable error and the maximum effective error, respectively. if $e_i \leq m$, we believe the current value of this indicator is exactly matched with the driver, and if $e_i \geq n$, we believe this indicator do not match the driver at all. However, the extreme pursuit of different indicators will lead to the non-convergence of training and get a bad result eventually, it is necessary to choose the appropriate $m$ and $n$. As for this paper, considering the range and precesion of each indicator, we choose $m=0,2$, $n=2$ for $t_f$, $t_{nf}$ and $m=0.5$, $n=5$ for $dv_{nb}$.

Finally, we can get reward functions for all three indicators, $r_f$, $r_{nf}$ and $r_{nb}$, and the total reward function $R$ is defined as

\begin{equation}
R = r_f + r_{nf} + r_{nb}.\label{eq13}
\end{equation}

\subsection{Neural network design and training details}\label{subsec5}

\begin{figure}
\centering
\includegraphics[width=0.4\textwidth]{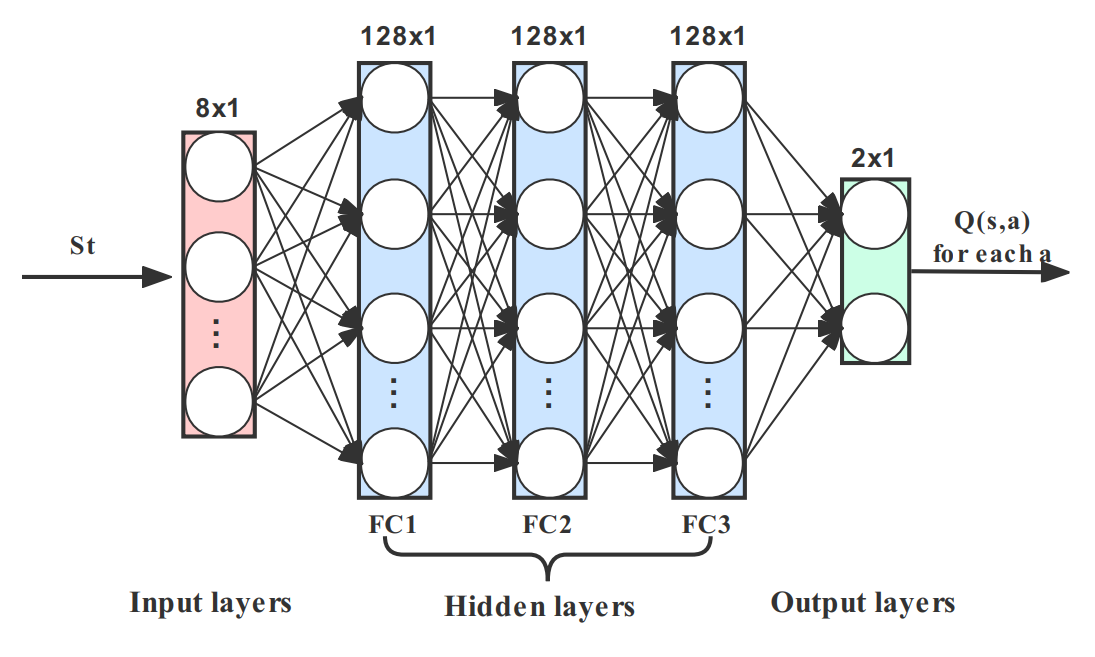}
\caption{The designed FCNN architecture}\label{fig8}
\end{figure}

A fully connected neural network (FCNN) architecture \cite{bib30} is designed for the target network and the prediction network mentioned in Section \ref{sec2}. As shown in Fig.\ref{fig8}, there are three hidden layers in this architecture, each layer has 128 neurons, and the activation function rectified linear units (ReLUs) are used. The input is a state vector of $8×1$ size, the output is a state-action value vector of $2×1$ size. At any simulation time $t$, the neural network gets an environment state $s_t$ and calculates the estimation state-action values $Q(s,a)$ for each action $a_i$ in action space $A$.

\begin{table}[h]
\begin{center}
\caption{Hyper parameters for training}\label{tab2}%
\begin{tabular}{lccc}
\hline

Symbol & Parameters  & Value \\ \hline

$\eta$    & Learning rate   & 0.005  \\
$\epsilon_s$    & Initial exploration   & 0.8  \\
$\epsilon_e$   & Final exploration   & 0.1  \\
$\gamma$	& Discount factor	& 0.98 \\
$M_r$  &Replay memory size	 &10000 \\
$M_i$  &Initial minimal memory size	&2000 \\
$M_m$  &Mini-batch size	&32 \\
$N_u$  &Target network update frequency	&20 \\
$N_s$  &Maximal episode step	&200 \\
$N_e$  &Maximal training episode	&10000 \\ \hline
\end{tabular}
\end{center}
\end{table}

The neural network is trained with a learning rate $\eta$ by using the DQN algorithm mentioned. The $\epsilon$-greedy policy is applied in training, and along with the training process, the value of $\epsilon$ will decrease from $\epsilon_s$ to $\epsilon_e$ linearly. The discount factor $\gamma$ is used to consider the discount of future reward. We set a replay memory with size of $M_r$, and the training begins after the initial minimal memory size $M_i$, and the random sample mini-batch size is set to $M_m$. The weights of target network $\theta^-$ is replaced by prediction network's weights $\theta$ every $N_u$ episodes. Every episode starts with a random environment state and stops with a TO CHANGE lane decision or maximal episode step $N_s$, and the maximal training episode is set to $N_e$.The hyper parameters used in training are summarized in Table \ref{tab2} with specific values.

\section{Results and discussion}\label{sec5}

In this paper, three personalized RL agents are designed and trained for lane change decision making to reproduce the typical drivers’ preferences. In the lane change scenario mentioned in Section \ref{sec4}, agents experience different states and learn to make decision themselves by repeating the lane change interaction, and finally, a stable policy for any states can be learned.

\subsection{Training results}\label{subsec2}

\begin{figure}
\centering
\includegraphics[width=0.5\textwidth]{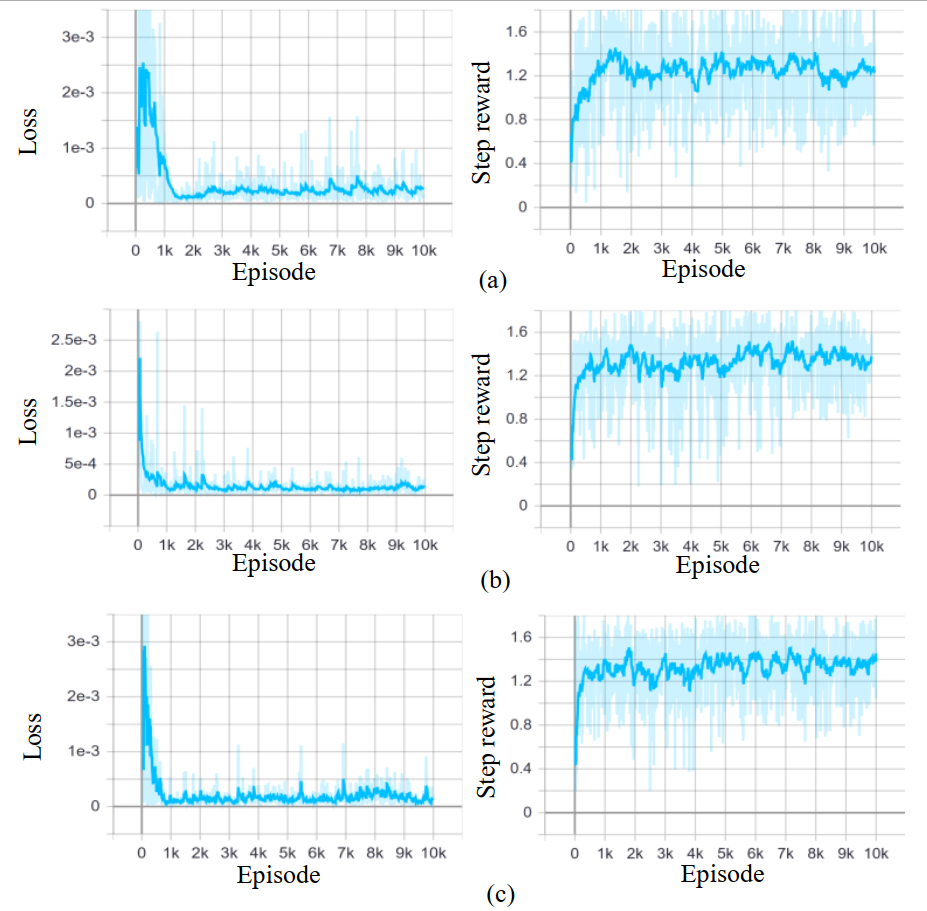}
\caption{Training losses (left column) and step rewards (right column) for personalized RL agents. The dark blue curves are obtained by smoothing the real values in light blue color. (a) Defensive agent. (b) Normal agent. (c) Aggressive agent.}\label{fig9}
\end{figure}

Fig.\ref{fig9} shows the training results of three RL agents with different lane change decision preferences. The horizontal axis is training episode, the training losses defined in (\ref{eq5}) are in the left column, and the step rewards defined in (\ref{eq13}) are in the right column. 
It is obvious that the loss curve has a quick downtrend in the first 1.5k training episodes and then flattens out, which means neural network converges after 1.5k episodes. As for the step reward curve, shows the average step reward is increased in every episode during the training process, implying that the RL agent learned to select the actions with higher reward in simulation a series of lane change maneuvers. Both loss and step reward eventually stabilize. Because of the random training environment, the curve of step reward is not very smooth but its mean value is stable at around 1.4.

\subsection{Comparison algorithm}\label{subsec3}

A comparison algorithm is set, in order to show the advantage of our proposed algorithm. With the designed reward function and the selected typical drivers, this algorithm directly compares the reward between actions TO CHANGE lane and NOT TO CHANGE lane at every step, and makes the decision with a higher reward. This is a kind of greedy strategy, and it will be deployed on three benchmark agents with Defensive, Normal and Aggressive styles. Further, they will be tested as comparisons in the same simulation environment together with the trained agents.

\subsection{Test and validation}\label{subsec4}

The trained RL agents and the benchmark agents are tested in our random simulation environment. They make lane change decision considering the environment states and the values of personalization indicators are recorded. Then three sets of lane-change points are obtained. The reference lines mentioned in Section \ref{sec2} and the lane-change points are compared, then the similarities are calculated by the Mean Absolute Error (MAE). To further illustrate the agent’s personalized preferences, the statistical results of decision making accuracy are presented through the comparison among the typical drivers, the trained RL agents and the benchmark agents.

\begin{figure}
\centering
\includegraphics[width=0.5\textwidth]{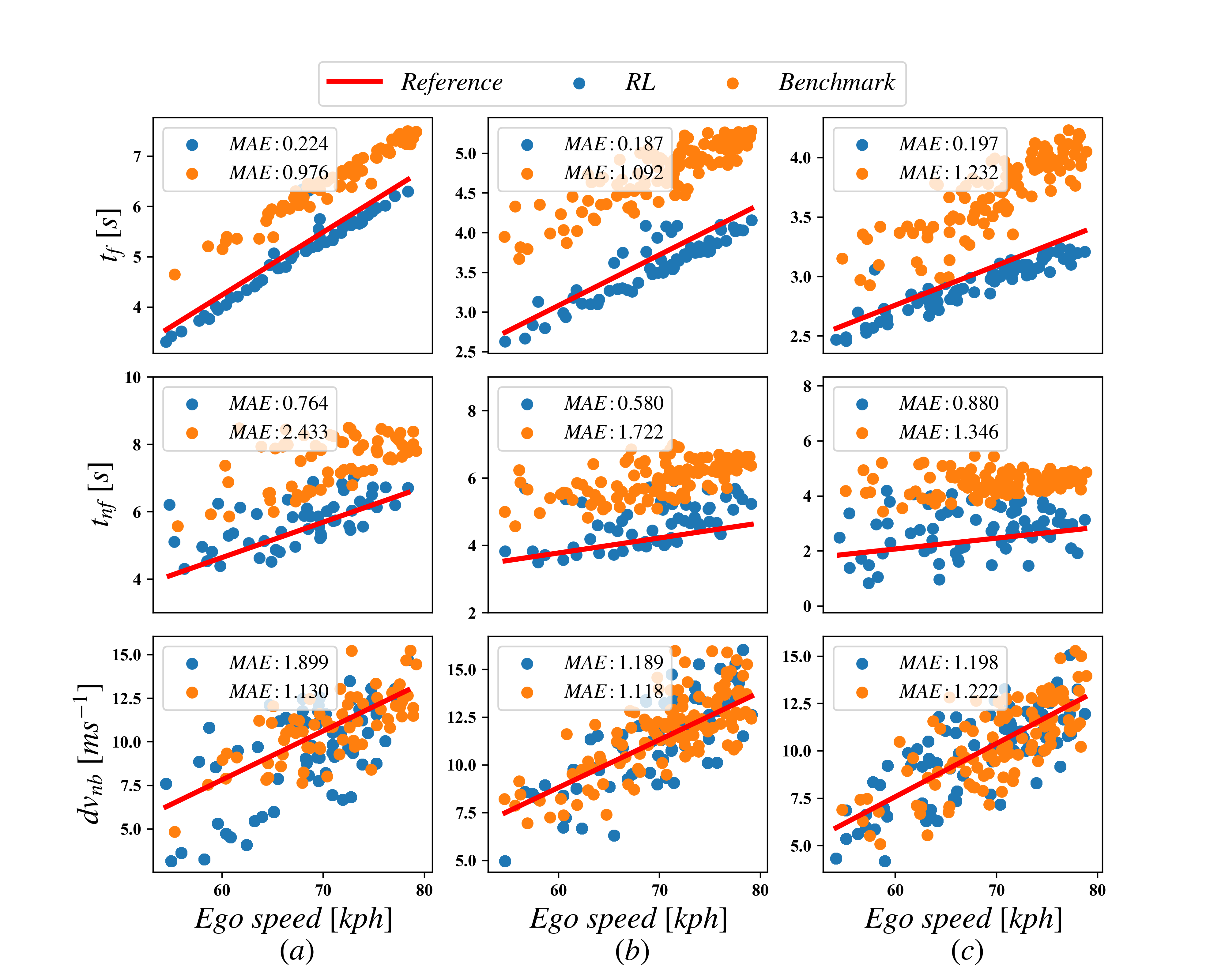}
\caption{Test results of personalization indicators in lane change decision at different ego speeds. (a) Defensive agents. (b) Normal agents. (c) Aggressive agents.}\label{fig10}
\end{figure}

Fig.\ref{fig10} shows the test results of three personalized RL agents and benchmark agents comparing to the typical drivers with different driving styles, i.e., Defensive, Normal, Aggressive. When the agents decide to make a lane change, the points in sub-plots represent the value of personalization indicators at different speeds of ego car, and the blue and the orange points are generated by RL agents and benchmark agents, respectively. 

To describe how similarly the agents and the corresponding drivers decide in lane change maneuvers, the similarities between reference lines and lane-change points are represented by the MAE, which is defined as 

\begin{equation}
MAE = \frac 1n \sum\limits_{i=0}^n \vert y_{a_i} - y_{r_i} \vert,\label{eq14}
\end{equation}
where $y_{a_i}$ is the actual value of indicators, and $y_{r_i}$ is the reference value that obtained from the reference line with the same speed as $y_{a_i}$.
As we can see, the three personalized RL agents, defensive, normal and aggressive, their lane-change points are close to the reference lines, that means when the agents making a lane change decision, the values of personalization indicators are close to the reference values. As for the benchmark agents, only the values of $dv_{nb}$ are close to the references, while the performance of the other two indicators, $t_f$, $t_{nf}$, is worse than the RL agents. Specifically, at the initiation of lane change decision, the MAEs of the RL agents of $t_f$, $t_{nf}$ are obviously lees than that of benchmark agents.

\begin{figure}
\centering
\includegraphics[width=0.5\textwidth]{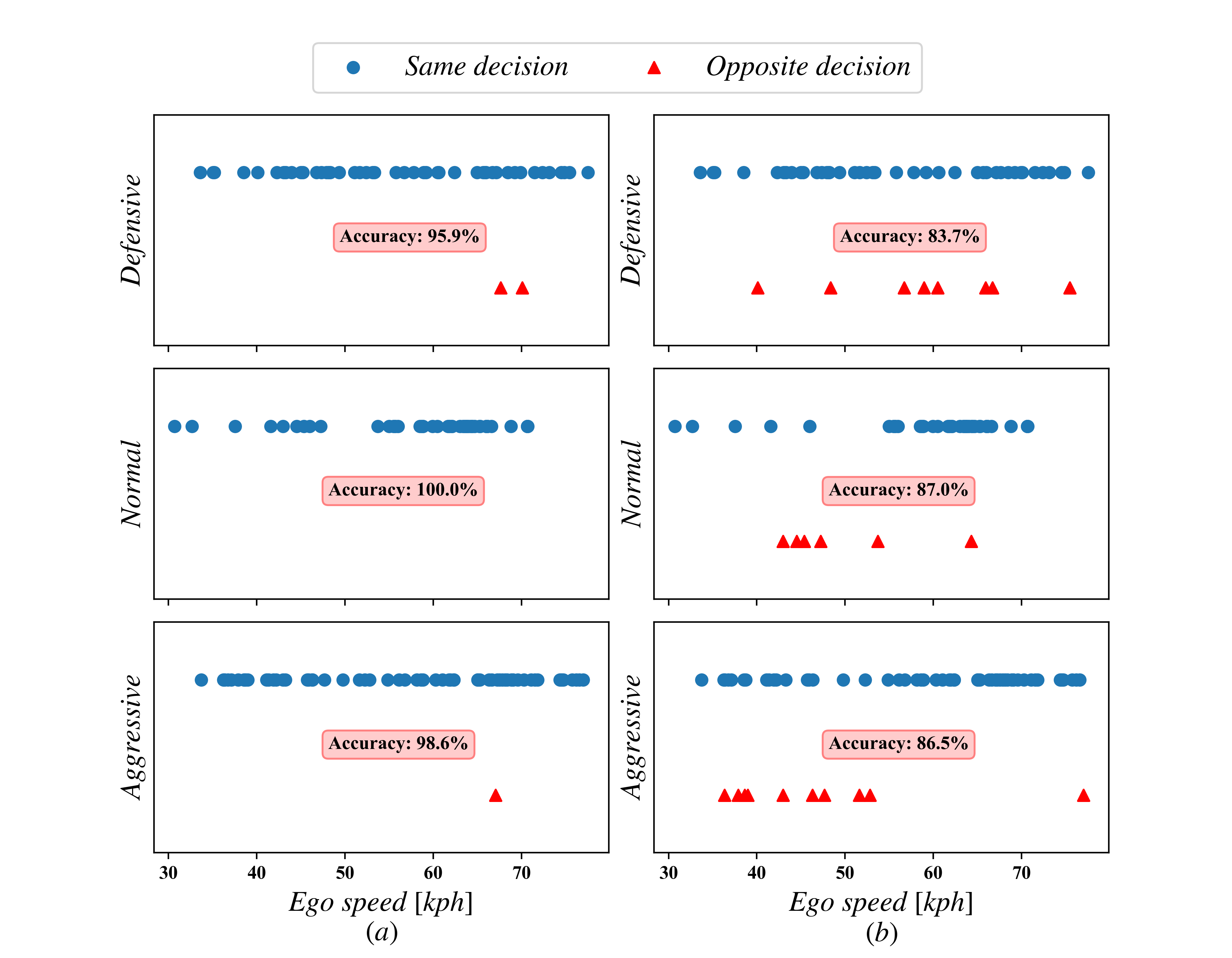}
\caption{The comparisons of lane change decision making results between human drivers, trained RL agents and benchmark agents with three different personalized preferences. (a) RL agents. (b) Benchmark agents.}\label{fig11}
\end{figure}

Further, the decision results of typical drivers, trained RL agents and benchmark agents in a series of same states are compared, with the statistical results shown in Fig.\ref{fig11}. The blue circle points represent that the human drivers and the agents make the same decisions, while the red triangles, in contrast, represent the opposite decision. For RL agents, we have 95.9\% accuracy for Defensive agent, 100\% accuracy for Normal agent and 98.6\% accuracy for Aggressive agent. However, the values of accuracy for benchmark agents are only 83.7\%, 87\% and 86.5\%, respectively.
For Defensive and Aggressive RL agents, there are totally three opposite cases marked in Fig. 11, comparing with original lane change data generated by typical drivers. To be specific, in cases 1 and 2, the relative distance ($d_{nb}$) between the ego car ($C_{ego}$) and the car behind in the target lane ($C_{nb}$) is too large, 142.36 meters and 126.11 meters, respectively. However, in our training environment, if $V_{nb}$ is too far away from the $C_{ego}$($d_{nb}>100$), we think $C_{nb}$ has no effect on the lane change decision making of $C_{ego}$. As for case 3, there is no leading car in the target lane ($C_{nf}$), so the related indicator $t_nf$ is recorded as -1 which contributes to a false result of decision, because our hypothesis is that $C_{nf}$ must exist when the agent makes lane change decision. All these opposite cases' environment states are not involved in our training environment, so the agent cannot handle them.  
To sum up, from the similarities of personalization indicators and the accuracies of decision making, and the analysis of opposite cases, we can tell that the personalized RL agents can make the lane change decision more like human drivers with three different lane change styles than the benchmark agents.

\section{Conclusions}\label{sec6}

Highway lane change is a kind of stressful and dynamic maneuver. To or not to change to the target lane at a certain speed, is a crucial but difficult decision task of lane change in dense traffic. This lane change decision problem, if further considering the personalized preferences of drivers for better user experience, is extraordinarily challenging in autonomous driving algorithm development. The related open questions include ``what preferences of human driver are important'', ``what different personalization indicators are'', and``How can the personalization style be learned'', etc.

Focusing on the lane change decision personalization, our main contributions can be summarized as follow.

(1) Three effective indicators for driver lane change decision preference description, TTC to the front car($t_f$), TTC to the front car in target lane($t_{nf}$) and relative velocity to the behind car in target lane($dv_{nb}$), are confirmed by the driver-in-the-loop experiments on the 6-DoF driving simulator. Driving data is collected and used to analyze driver preferences in lane change decision, and three typical drivers, specifically with Defensive, Normal and Aggressive driving styles are selected.

(2) Deep Q-Network reinforcement learning are viable for personalized lane change decision making algorithm. The aforementioned personalization indicators are used in reward function design to help the agents make a better decision in line with certain driver styles. The trained RL agents and the benchmark agents are tested in a two-lane highway scenario and are compared to human drivers with the same environment states. Results show that the RL agents can make lane change decision precisely like the human drivers with high decision accuracy, and perform better than greedy-policy based benchmark agents.

To conclude, by combining experiments of driving simulator and the human driver in traffic simulation, we acquire sufficient driver data and select three personalization indicators of different lane change decision. The newly proposed decision agent based on deep reinforcement learning can effectively reproduce the personalized styles of lane change decision and is promising for further application in human-centered design of autonomous driving algorithms.

In our study, based on driver data analysis we select only three personalization indicators of decision, which also intuitively accord with the driver risk perception during lane change maneuvers. But there may exist some other personalization indicators that affect lane change decision, especially if more variables of both the traffic and road/vehicle are considered. In future work, we will improve the agent performance by adding these inadequate considerations and will extend the RL approach to some other scenarios, i.e., the urban driving scenario.

\section*{Funding}
The author(s) disclosed receipt of the following financial support for the research, authorship, and/or publication of this article: This work was
supported by Department of Science and Technology of Zhejiang (No.2021C01SA601840, 2018C01058).

\bibliographystyle{unsrt}  
\bibliography{arxiv}

\end{document}